# Building an ASR Solution for Training and Assessing Children's Reading in Bambara

Yacouba Diarra
RobotsMali AI4D Laboratory
Bamako, Mali
research@robotsmali.org

Nouhoum Souleymane Coulibaly
RobotsMali AI4D Laboratory
Bamako, Mali
research@robotsmali.org

Mamadou Dembele
RobotsMali AI4D Laboratory
Bamako, Mali
research@robotsmali.org

Aymane Dembele
RobotsMali AI4D Laboratory
Bamako, Mali
research@robotsmali.org

Michael Leventhal
RobotsMali AI4D Laboratory
Bamako, Mali
research@robotsmali.org

***Abstract*—*Automatic speech recognition for children's reading remains underdeveloped for most African languages, including Bambara, despite its potential value for reproducible literacy assessment. We present An bɛ kalan, an open-source system for assessing children's reading in Bambara, developed through an end-to-end process linking field data collection, benchmark construction, model adaptation, a reading application, and classroom validation. A mobile collection and assessment app was used to collect 55 hours of raw reading speech from 60 children, from which we construct a public benchmark for Bambara child-reading assessment. Fine-tuning experiments compare Soloni, a Bambara-adapted Fast-Conformer ASR framework with TDT and CTC decoders, with QuartzNet, a compact convolutional ASR architecture. The best Soloni model reduces WER from 0.42 to 0.22 and CER from 0.15 to 0.08, substantially outperforming QuartzNet on the isolated benchmark. The experiments further show that repeated readings of the same texts provide architecture-dependent benefits: they substantially improve QuartzNet but add only marginal gains for Soloni, while SpecAugment regulates training without exceeding the best unaugmented configuration. Disaggregated analysis identifies children under 10 as the main source of residual errors, motivating targeted collection from younger readers. Ten classroom trials supported continued use of the application.*

***Keywords*—*automatic speech recognition; low-resource languages; Bambara; child speech; reading assessment; fine-tuning; public benchmarking; educational NLP*

***Data and code availability*—*Dataset/benchmark: huggingface.co/datasets/RobotsMali/an-be-kalan-bench.*

## I. INTRODUCTION

Automatic speech recognition systems have made considerable progress for resource-rich languages, but the vast majority of African languages, including Bambara, remain insufficiently equipped with annotated data, specialized models, and publicly available evaluation sets [1], [2]. This asymmetry is particularly pronounced in the educational field, where children's speech, characterized by developing acoustic profiles, irregular speech rates, and high fundamental frequencies, constitutes a distinct and challenging target domain, rarely represented in existing corpora [3], [4].

In Mali, structural constraints in the school system make systematic individual reading assessment difficult to implement at scale [5], [6]. Reading outcomes remain very low, with the World Bank–UIS estimating that 90% of Malian children at late-primary age are not proficient in reading [7]. An ASR-based assessment tool could support more scalable and reproducible reading assessment, provided it is adapted to the language, reading domain, and child population [8]–[10].

This article describes An bɛ kalan, an integrated and open-source effort to provide teachers with such a tool while producing resources reusable by the research community. Our contributions include collection of a dataset, models, a data collection and teaching app, benchmarks, and field evaluation. We adapt existing ASR models for Bambara to the domain of child reading in Bambara and evaluate them on a benchmark constructed for this purpose. We also examine the effect of SpecAugment, a standard data augmentation method for ASR [11]. This work connects the full pipeline from field data collection and benchmark design to model evaluation and classroom testing, allowing corpus characteristics, model performance, and user experience to be analyzed together.

## II. PRIOR WORK

Recent surveys of African ASR describe a field still constrained by limited annotated data, inconsistent benchmarking, weak reproducibility, and uneven coverage across languages and dialects [1]. Bambara has begun to receive more focused attention through datasets and benchmarks for spontaneous speech, radio speech, and formal read speech [2], [12], [13]. These studies establish the feasibility and importance of Bambara ASR, but they do not address the specific case of children reading aloud, where the target domain combines child speech, controlled reading prompts, educational vocabulary, and deployment constraints in schools.

Children's speech ASR is a distinct research area because children's acoustic and articulatory patterns differ from adult speech and often reduce recognition accuracy [3], [4]. This paper complements that literature by studying the problem in a low-resource African language rather than in the better-resourced languages that dominate children's ASR research.

ASR-based reading assessment has been studied as a way to produce scalable measures of oral reading accuracy and fluency [8]–[10], including recent work on oral reading fluency assessment in West Africa using large-scale speech models [14]. This paper extends prior ASR reading-assessment work from the evaluation of existing large models toward deployable offline ASR systems for a low-resource educational setting.

## III. CREATION OF THE CORPUS

### *A. Data Collection Protocol and Procedure*

A mobile application was used both to collect text-aligned reading recordings and to provide preliminary reading-assessment functionality. A data collection campaign was conducted in the Bamako region with five data collectors, who recorded children aged 9 to 18 reading aloud from 22 Bambara textbooks sourced from RobotsMali's GAIFE project [17].The protocol systematically rotated across source texts to avoid overrepresentation of any title and produce a more balanced training corpus.

### *B. Field-Based Adaptations*

Initial experiments in data collection from children indicated several adjustments in both interface design and back-end processing that increased the quantity of data usable for training from 71% to 92% [15]. These modifications included improved ergonomics for young users, phrase-by-phrase progression, optional audio prompts before reading, word-by-word feedback, replay before validation, background cloud synchronization, cumulative time tracking, segmentation heuristics for children's speech, and near-synchronous linguistic review to rapidly improve recording sessions.

### *C. Corpus Key Numbers*

The Bamako data collection campaign produced 55 hours of raw Bambara reading speech from 60 readers, with a concentration between 13 and 17 years old [16]. After cleaning, 47 hours were retained and 8 hours were discarded, corresponding to an 85% retention rate; removals were mainly due to erroneous session durations, incomplete readings, duplicates, title spelling variants, and inconsistent demographic metadata. The total source corpus without duplicates, referred to as the *main* dataset, consisted of 1.6 hours of unique reading material. The remaining 45.6 hours consisted of readings of texts already in *main* by different voices, the *duplicate* dataset. The estimated gender distribution was 35 girls (59%) and 25 boys (41%), based on metadata collected in the field. Each textbook was read between 19 and 31 times over a total of 521 sessions.

## IV. BENCHMARK AND EVALUATION DATASET

The models are trained and evaluated on RobotsMali/an-be-kalan-bench, using texts from the GAIFE collection [17]. The benchmark design is focused on child speech, regional acoustics, and domain-specific repetitive text structures [16].

The test set combines unique readings from 11 children across 11 books (53.4 minutes), with no overlap in content or readers with the other partitions. None of the books from the benchmark appear in any other partition, supporting a fair assessment of generalization. By design, the 22 textbooks from the data collection campaign and the benchmark books are separate in order to preserve the integrity of the assessment [16].

## V. EXPERIMENTAL FINE-TUNING PROTOCOL

Four experimental frameworks explored the interactions between vocabulary adaptation, data augmentation using SpecAugment [11], and acoustic diversity, comparing two architectures. All models are evaluated on the isolated test partition described in Section IV.

### *A. Experimental Configuration*

The experiments compare QuartzNet, an ultra-compact fully convolutional ASR architecture with approximately 18M parameters [18], with the Soloni model, a Bambara-adapted Fast-Conformer ASR model of approximately 114M parameters that combines a Time-and-Duration Transducer (TDT) decoder [19], [20] with a Connectionist Temporal Classification (CTC) decoding objective. Both models utilize pre-trained foundations developed on approximately 130 hours of open-source Bambara ASR corpora (jeli-asr [21] and African Next Voices [2], [22]), leveraging the base checkpoints *stt-bm-quartznet15x5-v2* and *soloni-114m-tdt-ctc-v2* from RobotsMali.

**Tokenizer & Vocabulary**: QuartzNet implements character-based CTC decoding utilizing a vocabulary size of 63 characters. Soloni incorporates dual independent decoders: a TDT network paired with a convolutional CTC decoder utilizing a SentencePiece Byte-Pair Encoding (BPE) tokenizer (vocab_size = 512).

**Hyperparameters & Optimization**: QuartzNet is optimized via NovoGrad (LR = 1e-3) with a CosineAnnealing scheduler over 100 maximum epochs. Soloni utilizes AdamW paired with a NoamAnnealing scheduler scaled at 1.5. Both pipelines used an 8% warmup ratio and early stopping based on validation-set WER. Training was performed on an NVIDIA A100 GPU with 80 GB of VRAM.

We structured our empirical analysis across four configuration variants to isolate the impacts of vocabulary constraints, data augmentation, and acoustic redundancy:

TABLE I. EXPERIMENTAL CONFIGURATIONS

| Experiment | Training Data | Augmentation Config | Objective |
|---|---|---|---|
| Exp1 | *main* (1.6h) | None | Evaluates token mapping performance given a highly constrained, pristine vocabulary. |
| Exp2 | *main* (1.6h) | SpecAugment (Freq: 4, Time: 10, Rect: 10) | Tests whether spectral masking forces generalization in the absence of physical acoustic variance. |
| Exp3 | *main* + *duplicate* (47.2h) | None | Measures whether dense multi-speaker variations act as regularizers or promote |

|  |  |  | overfitting to textbook vocabulary priors |
|---|---|---|---|
| Exp4 | *main + duplicate* (47.2h) | SpecAugment (Freq: 4, Time: 10, Rect: 10) | Examines the compound interaction between synthetic masking and diverse real-world acoustics. |

Note: The same four experiments were run for both baselines. QuartzNet uses a rectangular mask implementation (cutout), while Soloni uses the original SpecAugment implementation with time and frequency masking. The numbers in the configuration represent the maximum number of randomly applied masks.

The *main* and *duplicate* training split allowed us to test whether additional speaker and acoustic variation improves recognition when the textual content remains constrained. It also creates a risk that models learn repeated lexical patterns rather than more generalizable acoustic and linguistic representations.

**Training Execution**: All experiments were trained with the same stopping criterion within each pipeline. Training was monitored using validation-set WER, and early stopping was applied after 15 epochs without improvement. Training logs were retained for all configurations to compare convergence behavior across the *main*-only and *main + duplicate* conditions, with and without SpecAugment.

### *B. Architectural Rationale and Deployment Performance*

QuartzNet and Soloni were selected to support offline, low-latency ASR inference on low-specification mobile devices. This is an important deployment requirement in Mali, where internet use remains limited, geographically uneven, and constrained by affordability [23], [24]. Both architectures provide non-autoregressive CTC decoding paths, prioritizing fast local inference and implementation simplicity over the broader modeling capacity of large multilingual systems. This design choice also supports immediate feedback, which is important for ASR-based reading applications because low user-perceived latency and timely feedback are associated with greater learner attention [25], [26].

To assess the feasibility of local execution, we benchmarked inference on consumer-grade smartphones, specifically Tecno Camon 40 devices, using a non-optimized C-based runtime:

- **QuartzNet Inference**: Processes an 8-second audio segment in <300ms, offering a negligible resource footprint well-suited for legacy hardware architectures.
- **Soloni Inference**: Delivers complete CTC decoding within 800 to 900ms on an identical 8-second audio file.

Large multilingual ASR systems such as Whisper and MMS are important reference points for multilingual speech recognition, but they were not included in the present comparison because the study focuses on compact, fine-tunable architectures suitable for offline, low-latency inference on moderately priced smartphones. QuartzNet and Soloni were selected because they provide non-autoregressive CTC decoding paths and can be adapted within the data and deployment constraints of the target setting. The on-device benchmarks reported here therefore support the feasibility of local inference for the selected architectures, while evaluation of larger multilingual systems under matched device, latency, and fine-tuning conditions is left for future work. The resulting sub-second inference times are consistent with the design goal of providing timely feedback in an interactive reading application.

## VI. Model Results

### *A. Overall Performance*

Fig. 1 summarizes overall WER and CER across the evaluated model configurations.

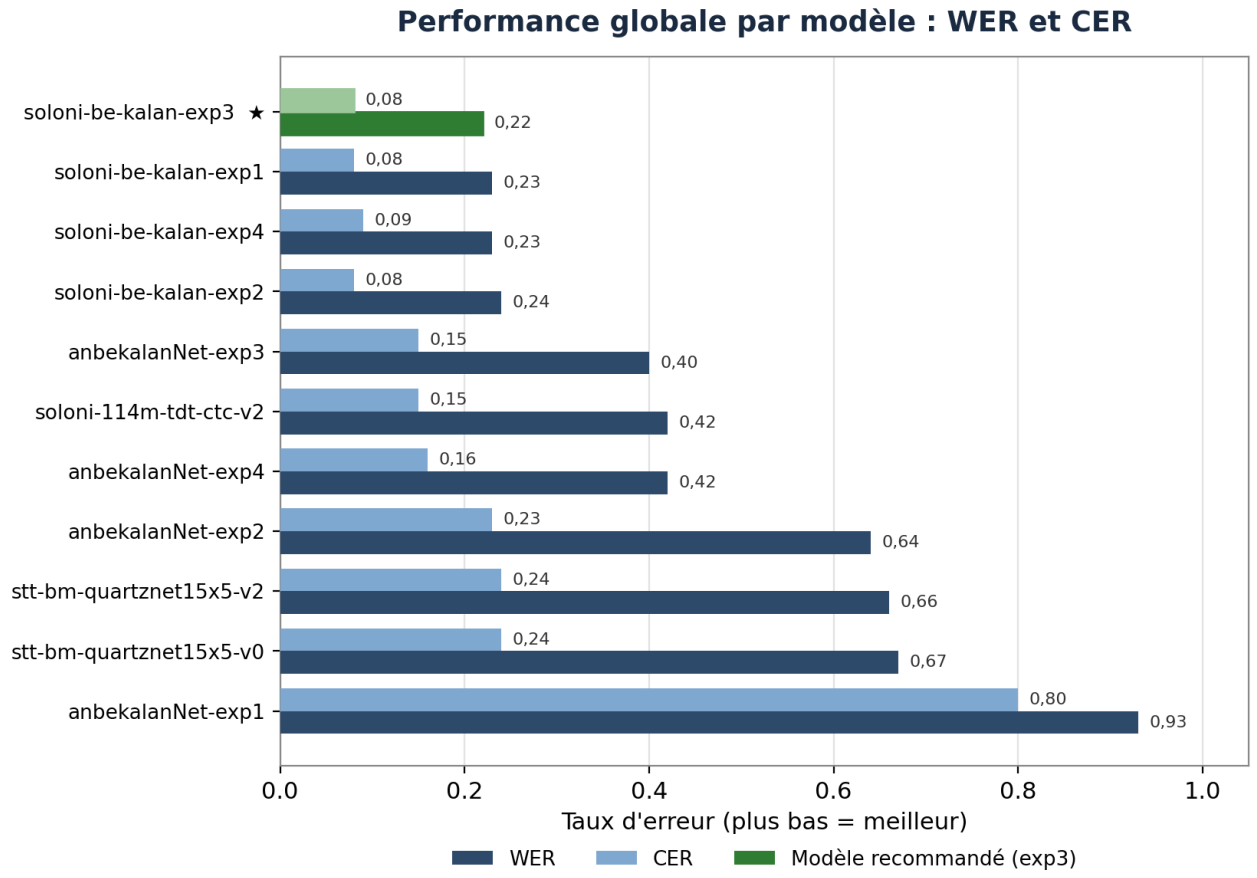

Fig. 1. Overall performance by model (WER and CER), sorted by ascending WER. The recommended model, soloni-be-kalan-exp3, is highlighted in green.

The recommended model, soloni-be-kalan-exp3, is the best performing configuration, with WER = 0.22 and CER = 0.08. It improves substantially on the Soloni base checkpoint, soloni-114m-tdt-ctc-v2 (WER = 0.42, CER = 0.15), and significantly outperforms the best QuartzNet configuration, anbekalanNet-exp4 (WER = 0.40, CER = 0.14). Within the Soloni family, Exp3 provides only a small WER gain over Exp1 (0.22 versus 0.23), while the SpecAugment configurations do not improve on the best unaugmented result: soloni-be-kalan-exp4 remains at approximately 0.23 WER and soloni-be-kalan-exp2 rises to 0.26 WER. By contrast, the QuartzNet configurations benefit strongly from the added repeated readings, falling from anbekalanNet-exp1 at 0.93 WER to anbekalanNet-exp3/exp4 at 0.40 WER, but they remain well behind Soloni. These patterns show that repeated-text acoustic variation is useful mainly for the weaker QuartzNet architecture, whereas Soloni's remaining constraint appears to be less raw acoustic volume than lexical and phonetic generalization to the isolated benchmark.

### B. Findings

- **SpecAugment as a passive regulator.** SpecAugment helped stabilize training and reduced divergence across the augmented configurations, but no augmented configuration outperformed the best unaugmented model. The unaugmented pipelines showed a large gap between training and validation behavior, with training-batch WER falling below 4% while validation WER plateaued at approximately 23% for Soloni and 40% for QuartzNet. This suggests that augmentation helped regulate optimization but did not add the lexical or phonetic diversity needed to improve generalization beyond the best unaugmented configuration.
- **Duplicate data mainly benefits weaker architectures.** Adding the *duplicate* subset substantially improved QuartzNet, reducing WER from 0.93 to 0.40. In contrast, the same additional data produced only a small gain for Soloni, from 0.23 to 0.22 WER. This suggests that the *duplicate* subset mostly reinforces repeated vocabulary and text patterns, which helps the weaker QuartzNet model avoid severe recognition failures but adds little new information for the stronger Soloni architecture [15].
- **Soloni's dominant constraint is lexical matching.** The quality and lexical diversity of the text take precedence over raw acoustic volume, directing the collection strategy toward phonetic and lexical richness rather than the mere accumulation of hours [15]. Both the Soloni and QuartzNet architectures were equally pretrained on the same amount of acoustic information prior to these experiments.

## VII. DISAGGREGATED ANALYSIS

### A. Residual Error Concentrated in Younger Readers

Fig. 2 shows WER by age cohort before and after fine-tuning.

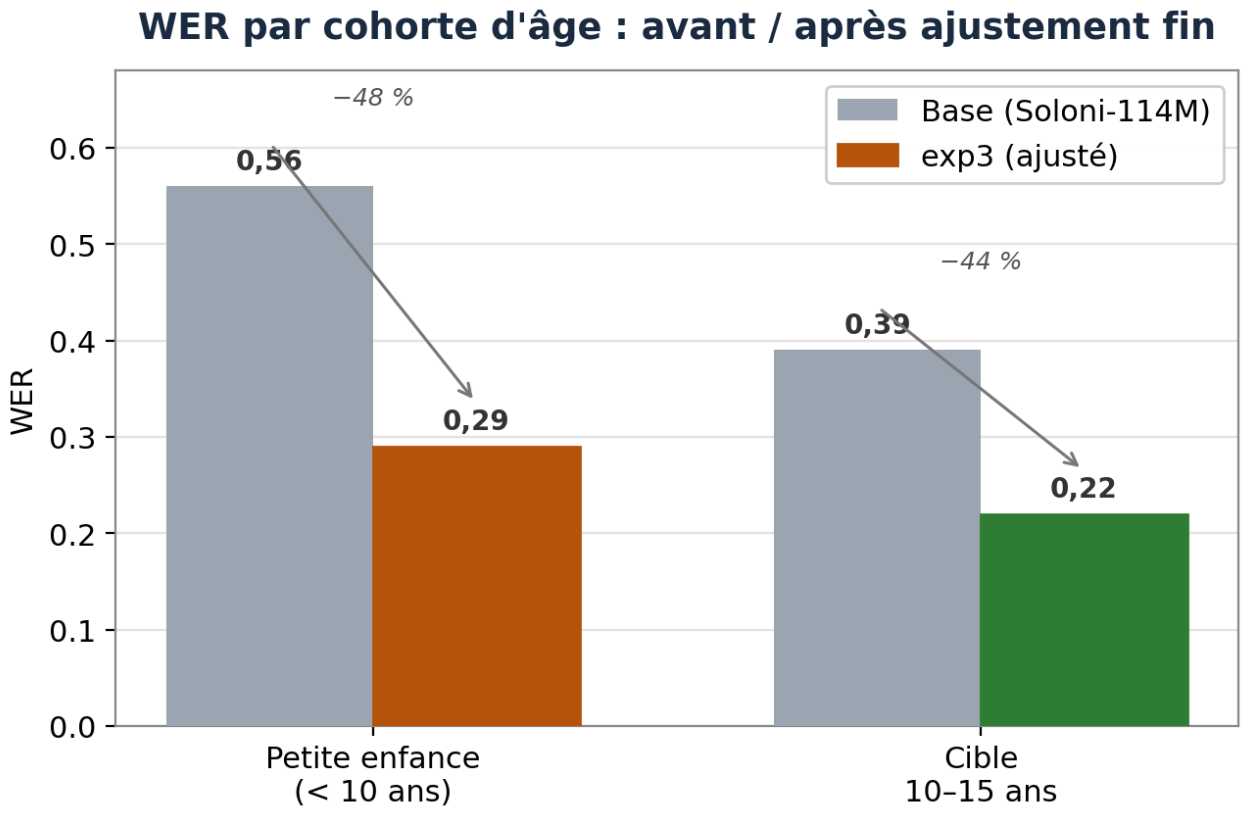

Fig. 2. WER by age cohort before (Soloni-114M base) and after fine adjustment (exp3). The orange bar indicates residual errors persisting in those under 10 years old (n = 93 statements; 527 for 10-15 year olds).

Children under 10 years old remain the primary source of errors across all models: the best model reduces their WER from 0.56 to 0.29, but this remains significantly higher than that of older cohorts. The acoustic and reading characteristics of early childhood, including less stable pronunciation, irregular speech rates, and higher fundamental frequencies, present a persistent out-of-domain challenge.

## VIII. USER TESTING

User testing was conducted in several schools in Bamako to evaluate the An bɛ kalan application under classroom conditions [15]. For each classroom trial, evaluators recorded teacher feedback and structured observations on the child's interaction with the application, the clarity and usefulness of the feedback, any technical or usability problems observed, and whether the trial supported continued use of the tool in classroom conditions.

All participating readers were aged 10 or older. The results indicate that the application was pedagogically acceptable in this initial field setting. In nine of the ten classroom trials, the application was judged suitable for continued use, although eight of these trials identified minor issues requiring correction. Teachers and children responded positively to the tool, especially to the word-level feedback and its use for Bambara reading practice.

## IX. DISCUSSION AND LIMITATIONS

The adapted Soloni model reduced WER from 0.42 to 0.22, substantially outperforming QuartzNet on the isolated benchmark, and should therefore serve as the reference baseline for subsequent work on Bambara child-reading ASR. This result, however, should be interpreted in relation to the deployment setting, since smaller architectures such as QuartzNet may remain relevant where memory, device cost, or inference constraints are dominant.

The results also show that increasing training hours through repeated readings of the same texts does not necessarily provide the same benefit as increasing linguistic and speaker diversity. Adding the duplicate subset substantially improved QuartzNet, reducing WER from 0.93 to 0.40, but produced only a marginal gain for Soloni, from 0.23 to 0.22. Similarly, SpecAugment helped regulate training but did not outperform the best unaugmented Soloni configuration. Taken together, these findings suggest that future improvements are more likely to come from targeted data collection than from simply adding repeated recordings, augmentation, larger models, or substantially greater compute.

An important remaining gap concerns younger readers. Children under 10 remain the main source of residual errors, and the benchmark contains only 93 statements from this cohort, compared with 527 from readers aged 10–15. Future data collection should therefore increase the representation of younger children while also expanding lexical and phonetic diversity through new reading material from the same educational domain. Field testing in ten classroom trials showed that the application was pedagogically acceptable, with nine trials supporting continued use.

ACKNOWLEDGMENT

This work was partially funded by the UNICEF Innovation Office (Stockholm), under contract no. 43456200. The authors thank the data collectors, the partner schools in Bamako, as well as the teachers and children who participated in the test sessions.